\pdfoutput=1

\documentclass{article}
\PassOptionsToPackage{numbers, compress, semicolon}{natbib}
\usepackage[final]{neurips_2021}

\usepackage[utf8]{inputenc} 
\usepackage[T1]{fontenc}    
\usepackage{hyperref}       
\usepackage{url}            
\usepackage{booktabs}       
\usepackage{amsfonts}       
\usepackage{nicefrac}       
\usepackage{microtype}      
\usepackage{xcolor}         

\usepackage{times}
\usepackage{latexsym}

\usepackage{comment}

\usepackage{amsmath}
\usepackage{amsthm}
\usepackage{amssymb}
\usepackage{bbm}
\usepackage{tikz-qtree}
\usepackage{tikz}
\usepackage{verbatim}
\usepackage{subcaption}
\usepackage{array}
\usepackage{inconsolata}

\usepackage [english]{babel}
\usepackage [autostyle, english = american]{csquotes}
\MakeOuterQuote{"}

\usetikzlibrary{tikzmark}
\usetikzlibrary{bayesnet}

\usepackage{cancel}

\usepackage{algorithm}
\usepackage{algpseudocode}
\algnewcommand{\parState}[1]{\State%
  \parbox[t]{\dimexpr\linewidth-\algmargin}{\strut #1\strut}}
\algrenewcommand{\algorithmiccomment}[1]{\hskip1em$\rightarrow$ \footnotesize#1 \normalsize}

\usepackage{float}
\usepackage{multirow}

\usepackage{microtype}

\usepackage[disable]{todonotes}
\makeatletter
\newcommand*\iftodonotes{\if@todonotes@disabled\expandafter\@secondoftwo\else\expandafter\@firstoftwo\fi}  
\makeatother

\newcommand{\note}[4][]{\todo[author=#2,color=#3,size=\scriptsize,fancyline,caption={},#1]{#4}} 

\newcommand{\jnote}[2][]{\note[#1]{josh}{white!40}{#2}}

\newcommand{\answer}[1]{``\MakeUppercase{#1}''}
\newcommand{\inputoutput}[1]{\textit{#1}}
\newcommand{\pipe}{$\mid$}

\usepackage{cleveref}
\crefname{section}{\S}{\S\S}
\Crefname{section}{\S}{\S\S}
\crefname{table}{Tab.}{Tabs.}
\crefname{figure}{Fig.}{}
\crefname{algorithm}{Algorithm}{}
\crefname{algorithm}{Algorithm}{}
\crefname{line}{Line}{}
\crefname{appendix}{App.}{}
\crefformat{section}{\S#2#1#3}
\crefname{thm}{Theorem}{}
\crefname{def}{Definition}{}
\crefname{prop}{Proposition}{}

\usepackage{booktabs, multirow} 
\usepackage{soul}

\definecolor{modelpink}{HTML}{D5A6BD}
\definecolor{modelgreen}{HTML}{b6d7a8}
\definecolor{modelorange}{HTML}{f9cb9c}

\definecolor{codec0}{HTML}{369275}
\definecolor{codec1}{HTML}{cc5d32}
\definecolor{codec2}{HTML}{5d709b}
\definecolor{codec3}{HTML}{c76aa3}
\definecolor{codec4}{HTML}{76a824}
\definecolor{codec5}{HTML}{bf990f}
\definecolor{codec6}{HTML}{e5c494}
\definecolor{codec7}{HTML}{b3b3b3}

\title{Decrypting Cryptic Crosswords: \\[0.5ex]
        Semantically 
       Complex Wordplay Puzzles as a Target for NLP}

\author{%
  Joshua Rozner \\
  Stanford University\\
  \texttt{rozner@stanford.edu} \\
  \And
   Christopher Potts \\
   Stanford University \\
   \texttt{cgpotts@stanford.edu} \\
   \And
   Kyle Mahowald \\
   University of Texas at Austin \\
   \texttt{mahowald@utexas.edu} \\
}

\date{}

\makeatletter
\patchcmd{\@verbatim}
  {\verbatim@font}
  {\verbatim@font\small}
  {}{}
\makeatother

\begin{document}
\maketitle

\begin{abstract}

Cryptic crosswords, the dominant crossword variety in the UK, are a promising target for advancing NLP systems that seek to process semantically complex, highly compositional language.
Cryptic clues read like fluent natural language but are adversarially composed of two parts: a definition and a wordplay cipher requiring character-level manipulations. 
Expert humans use creative intelligence to solve cryptics, flexibly combining linguistic, world, and domain knowledge. 
In this paper, we make two main contributions. 
First, we present a dataset of cryptic clues as a challenging new benchmark for NLP systems that seek to process compositional language in more creative, human-like ways.
After showing that three non-neural approaches and T5, a state-of-the-art neural language model, do not achieve good performance, we make our second main contribution: a novel curriculum approach, in which the model is first fine-tuned on related tasks such as unscrambling words.
We also introduce a challenging data split, examine the meta-linguistic capabilities of subword-tokenized models, and investigate model systematicity by perturbing the wordplay part of clues, showing that T5 exhibits behavior partially consistent with human solving strategies.
Although our curricular approach considerably improves on the T5 baseline, our best-performing model still fails to generalize to the extent that humans can.
Thus, cryptic crosswords remain an unsolved challenge for NLP systems and a potential source of future innovation.
\end{abstract}

\section{Introduction} \label{sec:introduction}

Modern computational models have made great progress at handling a variety of natural language tasks that require interpreting rich syntactic and semantic structures
\citep{devlin2018bert,radford2019language,manning2020emergent,rogers2020primer}. However, in NLP \citep{bender2020climbing,marcus2020next,bisk2020experience} as in other areas of AI \cite{lake2017building}, machines still lag humans on tasks that require flexible problem solving, rapid learning of unseen tasks, and generalization to new domains.
Just as complex games mastered by human experts, such as chess, Go, and video games, have proved a fertile ground for developing more flexible AI 
\citep{silver2018general,silver2016mastering,mnih2015human}, 
we propose that creative language games are a rich area for developing more flexible NLP models.
In particular, we argue that linguistic tasks involving meta-linguistic reasoning pose an important and significant challenge for state-of-the-art computational language systems.

One such domain is cryptic crossword puzzles. 
Cryptics are the most popular style of crossword in the United Kingdom and appear in major newspapers like \emph{The Times} and \emph{The Guardian}.
Cryptic clues have two parts: a \textbf{definition} and a \textbf{wordplay} cipher that, when placed adjacent to each other, read like fluent natural language.
For example, consider this NLP-centric cryptic crossword clue: ``But everything's really trivial, initially, for a transformer model (4)''
with answer \answer{BERT}.
``(4)'' is the \textbf{enumeration} and specifies the number of letters in the answer.
Solvers must identify which part of the clue is definition and which is wordplay. 
The \textbf{definition} should be a semantically valid description of the answer word: ``a transformer model'' can mean \answer{BERT}.
The \textbf{wordplay} part is the remainder of the clue: ``But everything's really trivial, initially''.
The word \textit{initially} in this context is used as an \textbf{indicator}, which tells the solver to take the \textit{initial letter} of each of the preceding words (\textbf{b}ut \textbf{e}verything's \textbf{r}eally \textbf{t}rivial) to produce \answer{BERT}.
Because both the wordplay and the definition give the same 4-letter word (which is what the \textbf{enumeration}, ``(4)'', calls for), we can be confident that \answer{BERT} is the correct answer.

\begin{table*}[t]
\small 
\centering
\caption{Examples of five common clue types in cryptic crosswords, all demonstrating clues for the answer: BERT. Indicators, where they occur, are italicized. The wordplay substrate appears in bold. Typographical emphasis added to aid reader, only; actual clues have no such indication.}
\begin{tabular}{p{0.35\linewidth} @{\hspace{20pt}}  p{0.2\linewidth} @{\hspace{20pt}} p{0.33\linewidth}}
\toprule
  \textbf{Clue type}  & \textbf{Clue example} & \textbf{Explanation for this example}\\
  \midrule
  \textbf{Anagram}: An anagram clue requires scrambling some set of clue letters to produce the answer. & 
  \textit{Confused}, \textbf{Bret} makes a language model (4) &
  \textit{Confused} indicates that we should ``confuse'' (anagram) the letters of ``Bret'' to get BERT. 
   \\ \midrule
  \textbf{Initialism}: An initialism requires taking the first letters of a phrase & \textbf{B}ut \textbf{e}verything's \textbf{r}eally \textbf{t}rivial, \textit{initially}, for a language model (4)  &
  \textit{initially} indicates taking the first letters of ``But everything's really trivial''. \\ \midrule
  \textbf{Hidden}: The answer occurs within a larger phrase.  & 
  Language model \textit{in} som\textbf{ber t}ext (4) & 
  \textit{in} indicates that a word is hidden inside a phrase.
  Extract the word BERT from the phrase ``somber text.''
  \\ \midrule
  \textbf{Charade}: For a charade clue, each part of the answer is clued sequentially. & 
  A language model exist? Right time! (4)  & 
  ``exist'' becomes ``BE'' since ``exist'' and ``be'' are synonyms. A standard abbreviation for ``right'' is ``R.'' A standard crossword abbreviation for ``time'' is ``T.'' 
  This clue type does not have an indicator.
  \\\midrule
 \textbf{Double definition}: In a double definition clue, two synonyms or phrases appear next to each other, each of which can refer to the answer. & 
 Model Sesame Street character (4)  & 
 Bert is a valid answer for "Sesame Street character", and it is also a model.
 Double definitions do not have indicators.\\ 
 \bottomrule 
\end{tabular}
\label{tab:clue_examples}
\end{table*}
The clue just discussed is an example of the \textit{initialism} clue type, which is one of roughly 15 major clue types.
Other types can require anagramming words, finding words hidden in longer phrases, performing synonym substitutions, substituting a word for a soundalike (e.g., "hiring" for "high ring"), or composing a number of these manipulations.
See Table \ref{tab:clue_examples} for examples of several other clue types 
with answer "BERT" and Appendix~\ref{app:clue_examples} for examples of actual clues from the dataset.
As with American-style clues, cryptics require world knowledge and linguistic flexibility to match the definition, but also considerable attention to meta-linguistic concepts to solve the wordplay. 
In this paper we study the language task of solving individual cryptic clues, rather than full puzzles since, unlike American-style crosswords, cryptics generally have unique answers and so are less reliant on grid constraints in order to achieve a complete solution.

While cryptics pose a challenge to novice solvers unfamiliar with their structure, experts flexibly combine world, domain-specific, and linguistic knowledge to solve novel clues.
Experts know the rules that govern cryptic clues, but they also reason about them flexibly and apply them to solve novel clues.
In the psychology literature, it has been claimed that cryptics depend on domain-general intelligence and are an ideal domain for studying the ``aha moment'' in humans \citep{friedlander2020fluid, friedlander2018penny}
because experts can solve them with high accuracy (but rarely at first glance), they can be easily generated, and they require drawing on a diverse range of cognitive abilities. 
Therefore, we believe that cryptic crosswords are an excellent domain for developing computational language systems that ``learn and think like humans'' \citep{lake2017building}, posing an interesting and important challenge for modern machine learning.

Our main contributions are, first, a cleaned dataset of cryptic crosswords clues from \texttt{theguardian.com}, consisting of 142,380 clues from 5,518 puzzles over 21 years;\footnote{We release the dataset along with all code to reproduce the results in this paper at \url{https://github.com/jsrozner/decrypt}.}
and second, a novel curriculum learning approach, in which the model is first fine-tuned on related, synthetic tasks (e.g., an augmented word descrambling task) before tackling actual cryptic clues. 
This method meaningfully improves on a standard T5 seq-to-seq approach and on the best model of \citet{efrat2021cryptonite}---concurrent work that presents a similar dataset and similar neural baseline using T5. 

In this paper, we aim not only to present a dataset and propose a novel solution but also to characterize the problem and motivate its importance.
To that end, we elucidate the task with three non-neural baselines and fine-tune T5 \citep{2020t5}, a Transformer-based \citep{Vaswani-etal:2017} encoder-decoder, as a neural baseline.
Since the character-level wordplay inherent to cryptics might be challenging to language models with subword tokenization (T5 uses SentencePiece \cite{kudo2018sentencepiece}),
we study whether T5 has or can acquire meta-linguistic knowledge.
In Section~\ref{analysis:length} we examine whether T5 learns meta-features of the task related to answer length.
In Section~\ref{analysis: descramble} we use a descrambling task to assess whether T5 understands the character composition of words and whether the model can make use of linguistic and meta-linguistic information simultaneously.
\jnote{we should have another experiment where we do just the definition with no letters, but maybe with answer length}
Our results show, perhaps surprisingly, that the subword-tokenized T5 model is quite robust to character-level challenges.
Moreover, the descrambling task may serve as a useful benchmark task in guiding the development of new approaches on the cryptics task.

Given the compositional nature of cryptic clues, we investigate the extent to which the model generalizes under increasingly difficult data splits.
In Section~\ref{subsec:disjoint} we introduce a new form of disjoint data split to address T5's robustness to inflection: a word-initial disjoint split that segments clue--answer pairs based on the first two letters of the answer.
In Section~\ref{analysis: systematic} we examine the systematicity of the model's answer generation by perturbing the wordplay portion of anagram clues, showing that its behavior is partially consistent with human solving strategies.

Although our novel curricular approach considerably improves performance on this task, fully solving cryptics remains a challenge for modern machine learning, with expert humans still outperforming machines.
Therefore, we hope that our dataset will serve as a challenging benchmark for future work.

\section{Related work}

While there is an existing literature on puns and wordplay in NLP \citep{punhe2019pun,punkao2013funny,punluo2019pun} as well as on solving American-style crosswords \citep{ginsberg2011dr,littman2002probabilistic,shazeer1999solving}, there has been relatively little work using NLP methods on cryptics, which require processing highly compositional language and disambiguating surface-level from hidden meaning.
\citet{hart1992cryptic} laid out a computational framework for solving the problem, and \citet{Hardcastle2001Using,hardcastle2007riddle} proposed some rule-based solutions. 
The app Crossword Genius from Unlikely AI solves cryptic clues and gives human-understandable explanations.
Because its method is proprietary and not available for open testing, we do not report it as a baseline but note that it is competitive.
\citet{deits_python} offers a rule-based solution, which can also output explanations, and which we test on our dataset.
Most recently, in work concurrent to ours, \citet{efrat2021cryptonite} release a similar dataset of cryptic crossword clues and present a neural baseline using T5-large. 
In Section~\ref{sec:comparison}, we discuss how our new data split and curricular approach improve on their work.

Our curricular approach fits into the space of recent work on pre-finetuning \cite{tafjord-etal-2019-quartz, aghajanyan2021muppet} and curricular approaches for compositional tasks \cite{geva-etal-2020-injecting, wu2021lime}.
Our approach is loosely related to the pre-finetuning of \citet{aghajanyan2021muppet} but differs in that our curriculum is composed of fewer tasks that are all closely related to the primary task and synthetically generated. 
Our approach resembles \citet{geva-etal-2020-injecting} in that we attempt to endow language models with a specific kind of reasoning by training in a multi-task setup over synthetic data.
In the vein of \citet{wu2021lime}, which trains on synthetic datasets to encode inductive bias, our approach can be understood as encoding wordplay functional biases. 

Whether large pretrained Transformer models generally pick up on meta-linguistic features like word length and the character composition of words is an open question.
\citet{brown-gpt3-neurips} explore a set of restricted word unscrambling tasks in the few-shot (versus fine-tuned) setting for GPT-3. 
Probing results from \citet{spellingbee-itzhak2021models} suggest that information about the character composition of words is present in the embeddings of pretrained, subword-tokenized models.
This seems to confirm our result that subword-tokenized models do have more knowledge of character composition and meta-properties of words than we might have expected.

\section{Dataset and task} \label{sec:Dataset}
We present a cleaned dataset of 142,380 cryptic crossword clues from 5,518 puzzles published in \emph{The Guardian} from July 1999 to October 2020.
We also introduce a challenging ``word-initial'' disjoint split after observing that T5 is robust to inflection.
Overall, the dataset has 55,783 unique answers, giving a mean frequency of 2.55 clues per unique answer. 
Answers in the dataset consist of one to six words, with most (97.4\%) having one (83.3\%) or two (14.1\%) words.
Full details of dataset preprocessing are given in Appendix \ref{app:dataset}, and in addition to releasing the full dataset, we include code to fully replicate our data download, pre-processing pipeline, and split generation in the repository.

\subsection{Task} \label{sec: data_task}
We frame the problem as a standard seq-to-seq task from inputs (clues with length enumeration) to outputs (answers). 
For example, one input could be \inputoutput{But everything's really trivial, initially, for a transformer model (4)}, 
with output \inputoutput{BERT}.
This is consistent with how clues are presented to human solvers.
Full details of the input--output specification for each model are provided in Appendix \ref{app:exp}, along with other experimental details.

\subsection{Splits} \label{sec: data_splits}

As motivated in the introduction (and in Section~\ref{subsec:disjoint}), we consider three splits. 
The \textbf{naive (random) split} is a standard 60/20/20 random split into train, dev, and test.
The \textbf{disjoint split} ensures that all clues with the same answer appear in only one of train, dev, or test. 
For example, if any clue for which the answer is \answer{BERT} is in the train set, then \textit{all} clues for which the answer is \answer{BERT} are in the train set. 
The disjoint split is used to test composition and generalization.
It prevents an approach that relies only on lexical similarity across clues (like KNN) from succeeding on the task.
Finally, the \textbf{word-initial disjoint split} is designed to address T5's robustness to inflection. 
For this split, we enforce that all clues whose answers start with the same two letters will appear in the same set. 
For example, all clues that have answers starting with `ab' like ``abacus,'' ``abdicate,'' ``abdicates,''  will be grouped, ensuring that inflections or spelling variations of the same base word occur in a single split.

\subsection{Metrics} \label{sec:metrics}
Our primary metric is whether the top-ranked output is correct, \textit{after} filtering to outputs of the correct length.
We filter because each clue in a puzzle has a hard length constraint, i.e., a human solver \textit{cannot} pencil in a solution that is of the wrong length.
Additionally, we report how often the correct answer is contained in the top 10 outputs after length filtering.
This is a useful metric since, when clues are presented in the context of a full puzzle, solvers use information from interlocking answers to narrow down a set of candidates.
For instance, the best-performing crossword solver for American-style (non-cryptic) crosswords relies heavily on satisfying grid constraints \citep{ginsberg2011dr}.
Comparing post-length-filter results from Section~\ref{baseline:results} with pre-filter results from Section~\ref{analysis:length}, the length filter is seen to increase the top-10 metric by roughly 6\% for T5 (with length enumeration given).

\section{Baseline models and results} \label{sec:baselines}
To characterize how simpler approaches perform on this task, we test three non-neural baselines: 
a WordNet-based heuristic model, a k-nearest-neighbor bag of words model (KNN BoW), and a rule-based model designed for the task \citep{deits_python}.
For a neural baseline, we fine-tune T5-base \citep{2020t5}.
For models that can interpret the answer-length enumeration as a textual token (KNN and T5), we append it to the input string (e.g., ``\texttt{(4)}'').
For these two models, we report results both with and without appending the enumeration.
Implementation details are discussed in Appendix  \ref{app:exp}.

\subsection{WordNet}
Our first baseline is a simple heuristic based on clue structure.
It takes advantage of the fact that the definition part of a cryptic clue always appears at the beginning or end of the clue. 
For instance, in the double definition clue for \answer{BERT} in Table~\ref{tab:clue_examples}, ``Model Sesame Street character,'' the word ``model'' appears at the beginning and is a definition (in this case, a hypernym) for the answer \answer{BERT}.
We use WordNet \citep{fellbaum1998wordnet}, a large database of English word meanings, to build a reverse dictionary
via the synonym, hyponym, and hypernym graphs.
We take as candidate solutions the set of reverse dictionary outputs for the first and last words of a clue.
For example, if ``dog'' appears at the start of a clue, candidates would include ``animal'', ``canine'', ``labrador'', etc.
Ranking outputs by character-level overlap with the rest of the clue slightly improves performance, since answers sometimes are rearrangements of the characters in the wordplay portion of the clue.

\subsection{KNN BoW}
To assess whether clues that are close in lexical space have similar answers, we use a KNN model on top of a bag-of-words featurizer. 

\subsection{Rule-based} \label{baseline:rule-based}
Finally, to evaluate how well a rule-based approach, with a hand-coded grammar, can perform on the task, we use the \citet{deits_python} solver.\footnote{\citet{deits_python} has a more recently implemented solver in Julia that was used by \citet{efrat2021cryptonite}. 
We use the Python version, which may have small differences from the Julia version and is reportedly much slower (see Appendix~\ref{app:exp} for more details).}
This solver handles anagrams, initialisms, substrings, insertions, and reversals in a rule-based way.
While the rule-based version includes common clue types, an inherent limitation of this approach is that it is difficult to enumerate all possibilities.
For instance, the Deits solver does not include charade-type clues, nor double definitions.
Moreover, the rule-based solver uses WordNet's \cite{fellbaum1998wordnet} word similarity functionality to rank outputs, meaning that, in general, it will fail on clues that have definitional parts consisting of more than one word (e.g. ``\texttt{language model}'' from our example in the introduction would not be matched).

\subsection{T5: vanilla seq2seq} \label{exp:vanilla}
For our baseline neural seq-to-seq approach, we fine-tune the Transformer-based \citep{Vaswani-etal:2017} T5-base model \citep{2020t5}, starting from HuggingFace's \citep{wolf-etal-2020-transformers} pretrained model parameters.
T5 is an encoder-decoder language model pretrained on the C4 corpus \cite{2020t5}. 
Fine-tuning is done via supervised learning (teacher-forcing) over standard seq-to-seq input-output pairs.
At test time, we generate outputs using beam search. 
As described in Section~\ref{sec:metrics}, we filter the outputs to those of the correct length and evaluate by checking whether the top result is correct and whether the answer occurs in the top 10. 
See Appendix ~\ref{app:exp} for details, including hyperparameter selection.

\begin{table*}[tp]    
\small
\centering
\caption{Results for baselines and top curricular approach. Details on the curricular approach are given in Section~\ref{sec:curr}. Metrics are percentages calculated over the top ten model outputs, after filtering to outputs of correct length.}\label{tab: Results_main}
\newcommand{\spacer}{\hspace{20pt}}
\setlength{\tabcolsep}{9pt}
\newcolumntype{L}{>{\raggedright}p{0.17\linewidth}}
\begin{tabular}
{@{} L *{4}{r} @{\spacer} *{4}{r} @{}}
\toprule
\multirow{3}{*}{\textbf{Model}} &  
    \multicolumn{4}{c@{\spacer}}{\textbf{Naive (random) split}} & 
    \multicolumn{4}{c}{\textbf{Word-initial disjoint split}} \\
    
&  \multicolumn{2}{c}{\textbf{Top correct}} & 
    \multicolumn{2}{c@{\spacer}}{\textbf{Top 10 contains}}  &  
  \multicolumn{2}{c}{\textbf{Top correct}} & 
    \multicolumn{2}{c}{\textbf{Top 10 contains}}   \\
\cmidrule{2-5}\cmidrule{6-9}
& {dev} & {test} & {dev} & {test} 
& {dev} & {test} & {dev} & {test} \\

\midrule
WordNet & 2.8 & 2.6 & 10.8 & 10.7 & 2.6 & 2.6 & 10.6 & 10.5 \\ 
Rule-based & 7.2 & 7.3 & 14.8 & 14.7 & 7.4 & 7.3 & 14.9 & 14.5 \\ 
\midrule
KNN (no lengths) & 5.6 & 5.6 & 9.9 & 10.1 & 0.0 & 0.0 & 0.0 & 0.0 \\ 
KNN (lengths) & 6.0 & 6.1 & 11.2 & 11.3 & 0.0 & 0.0 & 0.0 & 0.0 \\ 
\midrule
T5 (no lengths) & 15.3 & 15.6 & 29.4 & 30.0 & 0.9 & 1.0 & 4.8 & 5.1 \\ 
T5 (lengths) & 16.0 & 16.3 & 33.1 & 33.9 & 1.1 & 1.1 & 5.6 & 5.8 \\ 
\midrule
Curricular: ACW + ACW-descramble & \textbf{21.5} & \textbf{21.8} & \textbf{42.2} & \textbf{42.4} & \textbf{6.1} & \textbf{6.5} & \textbf{18.9} & \textbf{20.0} \\ 
\bottomrule
\end{tabular}
\end{table*}

\subsection{Results} \label{baseline:results}
In Table \ref{tab: Results_main}, we report metrics for dev and test sets on both the naive (random) split and word-initial disjoint split (discussion on disjointness in Section~\ref{subsec:disjoint}).
While the WordNet baseline achieves some success (2.6\% and 10.7\% top-1 and top-10 on the test set), it is inherently limited, since it cannot handle clues with multiword definitions and lacks a good ranking mechanism. 
KNN does better, achieving 6.1\% and 11.3\% with length enumeration.
The rule-based solver achieves 7.3\% and 14.7\%, marginally outperforming the KNN baseline.
Though our T5 neural baseline outperforms all non-neural baselines, achieving 16.3\% and 33.9\%, it leaves considerable room for improvement.

\section{Curriculum learning} \label{sec:curr}
Solving cryptic clues uses different linguistic and reasoning abilities compared to the natural language tasks on which T5 is pretrained. 
Thus, during fine-tuning on the cryptics task, the model must learn many sub-parts of the problem simultaneously: how to look up definitions, how to identify wordplay indicators, how to perform wordplay operations, etc. 
Although these elements of the task each individually benefit from T5's natural language pretraining, our standard T5 baseline suggests that learning to compose them all at once is challenging.
We show that a curriculum of synthetic datasets can address this, substantially improving performance on the primary cryptics task.
We test a number of different curricular tasks and task sets and discuss which are most effective.
This curricular approach may be useful in other settings where focused linguistic capabilities are required.

\subsection{Curricular datasets} \label{curr: datasets}
The logic of our curricular approach is to provide some guidance on the sub-parts of the problem space before building up to fully compositional cryptic clues.
For curricular training, we create four datasets designed to improve performance and elucidate what makes a good curricular task for this problem.
We process a public American crossword clue dataset \cite{xd_cw_paul}, henceforth ``ACW-data'' for ``American Crossword'', and generate three datasets: ACW, ACW-descramble, ACW-descramble-word. 
After preprocessing, ACW-data has 2.5m clue--answer pairs with 250k unique answers, giving a mean frequency of roughly ten clues per unique answer.
Unlike cryptic clues, American crossword clues often involve relatively straightforward synonym or definition substitions, so the ACW dataset can be used to train definition lookup. 
We also produce a separate anagram dataset from a publicly available English dictionary. 
Details to produce all datasets are included in Appendix~\ref{app: curr: datasets}. 
In all example clues that follow, the target output is \answer{petal} and we use labels (prepending a word and colon) to help the model distinguish tasks.
\begin{enumerate}
\item 
\textbf{ACW}: ACW-data dataset in input--output form (i.e., a seq-to-seq version of American-style crossword clues) with no augmentation.
For example, \inputoutput{phrase: flower part (5)}.
\item 
\textbf{ACW-descramble}: For each clue--answer pair in ACW-data, we create an input that models a cryptic clue by scrambling the answer word and prepending or appending it to the clue portion.
For example, we scramble ``petal'' and randomly place it at the beginning (\inputoutput{descramble: etalp flower part (5)}) or end (\inputoutput{descramble: flower part etalp (5)}) of a clue.
\item
\textbf{ACW-descramble-word}: A seq-to-seq task that is just the descrambling of answer words. When compared to ACW-descramble, this elucidates the importance of curricular and primary task similarity, in particular whether teaching a bare descrambling task is helpful to the model. Example: \inputoutput{descramble word: etalp (5)}.
\item
\textbf{Anagrams}: Using a dictionary as starting point, we synthesize an anagram dataset:
we pair a word (to be anagrammed) with an anagram indicator (e.g., "mixed up", "drunk", "confusingly") and ask the model to produce a valid anagram (i.e., a scrambled version of the word that is itself also a valid word).
For example, \inputoutput{anagram: confusingly plate (5)} (rearrange the letters of `plate' to get \answer{petal}). 
The anagram dataset simulates the anagram type of wordplay in cryptic clues, with definition part omitted.
\end{enumerate}

\subsection{Methods}

Our curricular approach is as follows:
using the same methods as in Section~\ref{exp:vanilla},
first, we fine-tune T5 on one or more supplementary tasks.
Second, we continue fine-tuning on the primary cryptics task, periodically showing batches from the supplementary task(s), to decrease the likelihood of catastrophic forgetting \citep{FRENCH1999128_catastrophic}. 
Full details of our curricular training setup are in Appendix~\ref{app: curr: training}.

\begin{table*}[tp]     
    \small
    \caption{Curricular results (left) and sample metrics for meta-linguistic feature analysis (right)}
    \begin{subtable}[t]{0.42\linewidth}
    \centering
    \caption{Curricular approaches on the naive (random) split. Metric is correctness of top-output (5~beams with length filter).} \label{tab: curricular}

    \newcolumntype{R}{>{\raggedleft\arraybackslash}p{1cm}}

    \setlength{\tabcolsep}{2pt}
    \begin{tabular}[t]
    {@{} p{0.6\linewidth} r r @{}}
    \toprule
     & \multicolumn{2}{c}{Percent correct} \\
    Curricular dataset &  
    \multicolumn{1}{R}{full dev set} & \multicolumn{1}{R}{anagram subset} \\
    \midrule
    Baseline (no curricular) & 15.7 & 13.7\\
    \midrule
    ACW & 18.3 & 14.4 \\
    ACW-descramble & 13.1 & 21.4\\
    \midrule
    ACW + ACW-descramble & \textbf{20.2} & 24.0 \\
    ACW + ACW-descramble-word & 17.8 & 18.3 \\
    \midrule
    ACW + anagram & 17.1 & 19.1 \\
    ACW + ACW-descramble + anagram  & 20.1 & \textbf{27.1} \\
    \bottomrule
    \end{tabular}
    \end{subtable}%
    \hfill
    \begin{subtable}[t]{0.55\linewidth}
    \caption{Sample metrics calculated over top 10 outputs \textit{without} length filter, using naive split.\\} \label{tab: sample features}
    \footnotesize
    \centering
    \newcolumntype{S}{>{\centering\arraybackslash}p{1.5cm}}
    \setlength{\tabcolsep}{3pt}
    \begin{tabular}[t]
    {@{} l *{6}{p{0.75cm} } @{}}
    
    \toprule
    
    \multirow{2}{*}{Model} &  
        \multicolumn{2}{S}{\% sample contains answer (top-10, no filter)} &
        \multicolumn{2}{S}{\% outputs with correct length} &
        \multicolumn{2}{S}{\% outputs correct word count}
        \\
    \cmidrule{2-7}
    & dev & test & dev & test &dev & test\\
    \midrule
    KNN \\
    -- (no lengths) & 6.5 & 6.6 & 13.4 & 13.3 & 70.7 & 70.7 \\ 
    -- (lengths) & 10.6 & 10.7 & 85.4 & 85.3 & 89.7 & 89.6 \\ 
    \midrule
    T5-base \\
    -- (no lengths) & 19.0 & 18.8 & 16.0 & 16.2 & 74.2 & 74.1 \\ 
    -- (lengths) & 27.5 & 28.1 & 48.3 & 48.5 & 97.9 & 97.9 \\ 
    \bottomrule
    \end{tabular}
    \end{subtable}
\end{table*}

\subsection{Results}

In Table \ref{tab: curricular}, we report results for our curricular approach. 
Overall, we find that curricular training using ACW + ACW-descramble is best and report in Table~\ref{tab: Results_main} a primary task improvement from 16.3\% to 21.8\% on the random split and from 1.1\% to 6.5\% on the word-initial disjoint split.

We begin by testing a curriculum with only ACW, which corresponds roughly to teaching the definitional lookup.
Observing that ACW improves performance over the baseline, we test ACW-descramble, which is ACW augmented with a scrambled word component (an implicit wordplay).
Surprisingly ACW-descramble leads to a decline in performance relative to both ACW and to Baseline. 
On the other hand, combining ACW + ACW-descramble leads to our best performing approach, demonstrating that a particular curricular combination can improve over two separate tasks.
We also compare ACW + ACW-descramble to ACW + ACW-descramble-word.
The drop in performance suggests that inputs that are more similar to the final task are more useful than, e.g., a bare descrambling task.
To isolate the effect of task choice, all curricular approaches use the same data distribution, train for the same number of curricular fine-tuning steps, and are reviewed during primary training at the same frequency.

Finally, in order to explore whether the curricular training improves performance across the board or just on the clue types directly involved (e.g., anagrams), 
we report performance on an algorithmically-labeled anagram subset ("anagram subset" column in Table~\ref{tab: curricular}). 
Adding the Anagrams subtask to the curriculum improves anagram performance but interestingly does not improve performance compared to the top curriculum, ACW + ACW-descramble.\footnote{The distribution and size of the Anagrams dataset is different from the ACW datasets, so we separate curricula involving the Anagrams dataset in Table~\ref{tab: curricular}.}
We see a similar gain in anagram performance (but drop in overall performance) when training with only ACW-descramble.
This suggests that pretraining for a particular clue type can improve performance on that type, but perhaps at the expense of performance on other clue types. 

Beyond providing guidance to T5 on the problem sub-parts,
this approach also partially addresses the disjointness of train and test sets. 
For the word-initial disjoint split, by periodically refreshing the curricular task, we remind T5 to produce outputs that are not in the training set of the cryptic split.
\jnote{jesse makes a good comment that we should isolate the effect of distribution vs generalization learning. add comment about the future work / figuring out better compositional possibilities?, see commented language below}

\section{Model analysis} \label{sec: model analysis}

\subsection{Learning meta-linguistic properties: output length and number of words} \label{analysis:length}

This task requires not just linguistic but also \textit{meta-linguistic} reasoning.
We investigate how model behavior changes when the enumeration (the specification of the length of the answer) is appended to the end of the input string as a number in parentheses. 
Whether large pretrained transformer models generally pick up on meta-linguistic features like word length is an open question. 

To study whether the models learn length information, we report how often the top-10 candidate outputs for each clue are the correct length and have the correct number of words \textit{before} applying any length filter.
For both the KNN and the T5 models, we find that including the length enumeration improves overall performance, as can be seen in Table~\ref{tab: Results_main}.
(We omit the WordNet and rule-based approaches from this discussion since they have no capacity to learn the meaning of the length enumeration.)

In columns 3 and 4 of Table~\ref{tab: sample features} we see that both KNN and T5 pick up on length information, generating more outputs of the correct length when the enumeration is provided. 
T5 is particularly proficient at producing the correct number of words in outputs.
(Recall that multiword answers are indicated with an enumeration that has two numbers separated by a comma, as in ``(4, 6)'', indicating an answer like ``ALAN TURING''.) 
Given that T5 produces 97.9\% of outputs with the correct number of words, 
it seems plausible that the model is learning a correspondence between the enumeration and the presence of a space or spaces in its output.

\subsection{Disjointness} \label{subsec:disjoint}

Based on the performance of the KNN model on the naive data split, we see that some clues can be solved by picking up on lexical similarity to other clues. 
Thus, we investigate whether T5 is also picking up on similarity to previously seen clues or if it is learning something about the compositional and functional structure of the cryptic clues.

To assess how much a Transformer-based model like T5 relies on having seen similar clues for the same word, we segment performance on the random split by whether the answer was in the train set.
In Table~\ref{tab: disjointness}, we see that performance drops from 16\% on the full dev set to only 3.0\% on the clue subset not seen during training, confirming our intuition that lexical similarity between clues with the same answer plays a role in model success.

To formalize this result, we create and train on the two disjoint datasets described in Section \ref{sec:Dataset}: the basic disjoint and the word-initial disjoint splits.
The T5 model achieves 3.3\% accuracy on the basic disjoint split (dev) and only 1.1\% accuracy on the word-initial disjoint split.
The drop is likely partially attributable to robustness to inflection, since inflected forms often start with the same two letters.

\subsection{Wordplay: minimal task} \label{analysis: descramble}
\begin{table*}[tp]     
    \caption{Disjointness results (left) and descrambling results (right)}
    \begin{subtable}[t]{0.53\linewidth}
    \caption{T5-base performance (\% for top-10 outputs after length filter) on naive, subset of naive not seen in train, disjoint, and word-initial disjoint splits.}\label{tab: disjointness}
    \small
    \centering
    \newcommand{\spacer}{\hspace{10pt}}
    \setlength{\tabcolsep}{4pt}
    \begin{tabular}
    { p{0.4\linewidth} r r @{\spacer} r r }
    \toprule
    
    \multirow{3}{*}{\textbf{Dataset}} &  
    \multicolumn{2}{c@{\spacer}}{\textbf{Top}} & 
    \multicolumn{2}{c}{\textbf{Top 10}} \\
    & \multicolumn{2}{c@{\spacer}}{\textbf{correct}} & 
    \multicolumn{2}{c}{\textbf{contains}} \\
    \cmidrule{2-5}
    & dev & test & dev & test \\
    \midrule
    Naive (random) split\\
    -- Entire split & 16.0 & 16.3 & 33.1 & 33.9 \\ 
    -- Subset not in train & 3.0 & 2.8 & 9.5 & 9.7 \\ 
    \midrule
    Disjoint splits: \\
    -- Naive disjoint
    & 3.3 & 3.2 & 12.6 & 12.9 \\ 
    -- Word-initial disjoint & 1.1 & 1.1 & 5.6 & 5.8 \\ 
    \bottomrule
    \end{tabular}
    \end{subtable}%
    \hfill
    \begin{subtable}[t]{0.44\linewidth}
    \caption{Descrambling task, with and without phrasal definition. Metric is \% for top ten outputs without length filter.}\label{tab: descrambling}
    \small
    \setlength{\tabcolsep}{3pt}
    \begin{tabular}
    {@{} l@{} c c @{}}
    \toprule
    \textbf{Split and task} & \textbf{Top} & \textbf{Top 10}\\
    & \textbf{correct} & \textbf{contains} \\
    \midrule
    Random split \\
    -- Descramble & 63.8 & 91.6 \\ 
    -- Descramble w/ phrase & 79.4 & 91.2 \\ 
    \midrule
    Word-initial disjoint split \\ 
    -- Descramble & 3.7 & 12.5 \\ 
    -- Descramble w/ phrase & 12.4 & 24.4 \\ 
    \bottomrule
    \end{tabular}
    \end{subtable}
\end{table*}

We investigate the extent to which T5, which uses SentencePiece tokenization \citep{kudo2018sentencepiece}, can perform wordplay-esque tasks like descrambling. 
We run an experiment as follows:
we start with the ACW dataset from Section~\ref{sec:curr}, further restrict to outputs with targets in a dictionary (i.e., no multiword answers), and downsample to 180k clues (10\%).
We create two descrambling tasks.
The first is a direct descramble task, where the input is a scrambled version of the target word (e.g., \inputoutput{etalp} for target \inputoutput{petal}).
The second task is a descramble with phrase tasks, in which we append the clue of the clue-answer pair after the scrambled answer letters (e.g., input is \inputoutput{etalp $\mid$ flower part} for target \inputoutput{petal}).
The second task is designed to mimic the cryptic setup, where we have a wordplay (in this case, the implicit descramble function) whose output is conditioned on a (possibly phrasal) synonym. 
See Appendix \ref{app: exp: descramble} for more details.

We present results in Table \ref{tab: descrambling}. 
We observe that the model reasonably learns the task on a random split (63.8\%) but fails on a word-initial disjoint split (3.7\%).
Notably, including a phrasal definition alongside the scrambled letters in the input improves outcomes, suggesting that the model is simultaneously incorporating both meta-linguistic character-level and overall word-embedding information. 
This task can serve as a benchmark for models to solve the cryptic task, since it roughly upper-bounds how well a model can solve wordplays.
\jnote{mention human parallels and how phrasal inclusion changes results}

Given that we observe a considerable drop from the random to disjoint data splits, we test whether T5 can learn the identity function under a disjoint split.
We find that the model achieves 100\% accuracy on a direct copy task in which the model is trained to simply output the string that is given as input.
This suggests that T5's meager performance for descrambling on the disjoint split is not due to an inability to generate an output that has not been seen in fine-tuning.

\subsection{Assessing systematicity} \label{analysis: systematic}
We investigate the extent to which the model's behavior is consistent with the compositional and systematic nature of human solving strategies. 
Consider our anagram clue from Table \ref{tab:clue_examples}: "Confused, Bret makes a language model (4)," for which the answer is \answer{BERT}.
Using a publicly available list of 18,000 first names, we algorithmically identify clues in the dev set that require the solver to compute an anagram of a first name, and we use those clues to compose two specialized test sets.
In the first set, we scramble the original letters (e.g., \inputoutput{Bret} becomes \inputoutput{Treb}).
If the model is behaving optimally, performance on this set should not suffer: what is important about \inputoutput{Bret} in this clue is not its semantics but its characters.
In the second set we substitute another name of the same length (e.g., \inputoutput{John} for \inputoutput{Bret}).
We pick first names because swapping a name does not change the grammaticality or linguistic validity of clues. 
Here, for a human solver, we  would expect a major hit in performance since the correct answer is a valid anagram of \inputoutput{Bret} but not of \inputoutput{John}, and so the clue is no longer a valid cryptic clue.
It does, however, still have a valid definition part ("a language model"), and so, if the model is picking up only on the definition part and ignoring the wordplay part, it might still do well.
See Appendix~\ref{app: wordplay} for more task details.

On this set of clues, prior to any modification, we find a baseline accuracy of 33.3\%.
When scrambling the name, accuracy drops moderately, to 25.2\%.
When substituting a name with different letters, accuracy falls to 7.0\%.
This difference in performance on the substitution and scrambling tasks suggests that the model is, to some extent, correctly picking up on the need to use the letters of the name as the wordplay substrate.
This is confirmed by the average character overlap between the altered word and the generated candidate answers.
We observe 51.2\% multiset overlap for the baseline (name unchanged from original clue), 51.0\% for substitution of names with the same letters, and 31.4\% when substituting names with different letters.
For our example clue, this means that we would expect high character overlap between \inputoutput{Bret} and the candidate answers in the baseline set, but high overlap between \inputoutput{John} and the candidate answers in the substitution test set.
These results suggest that, like human solvers, the model is sensitive to character manipulations at the location of the wordplay substrate.

\begin{table*}[tp]
\small
\caption{Performance of T5-large as reported by \citet{efrat2021cryptonite}, in our replication of their work, and with our top curricular approach (ACW + ACW-descramble). Metric is correctness of top output (5 beams without length filter) on test set.}\label{tab: Efrat Comparison}
\centering
\setlength{\tabcolsep}{4pt}
\begin{tabular}
{l r r r}
\toprule
\textbf{Split} &
\textbf{Efrat et al} &
\textbf{Our replication of Efrat et al} &
\textbf{Top curricular} \\
\midrule
Efrat `naive' (test) &
56.2 & 53.2 & 52.1  \\ 
Efrat `official' (test) & 
7.6 & 10.9 & \textbf{19.9} \\ 
Word-initial disjoint (test) &
-- & 4.9 & \textbf{12.8}\\ 
\bottomrule
\end{tabular}
\end{table*}

\subsection{Comparison to Efrat et al.} \label{sec:comparison}

In contemporaneous work, \citet{efrat2021cryptonite} present a dataset of cryptics from two other major newspapers and fine-tune T5-large for the task.
While \citet{efrat2021cryptonite} conclude that train/test disjointness is important, they do not fully consider T5's robustness to plural and other inflections. 
The word-initial disjoint split that we present addresses this.
In particular, their `naive' split is the same as our naive split, and their `official' split is the same as our (naive) disjoint split.
To demonstrate that our split is relevant to the Efrat work, we replicate their results (we train T5-large and report the same metric, correctness of top output with b=5 beams), 
show a considerable decline in performance under the word-initial disjoint split (10.9\% to 4.9\%),
and finally demonstrate that our curricular approach substantially improves results on the `naive-disjoint' (10.9\% to 19.9\%) and word-initial disjoint splits (4.9\% to 12.8\%).
Performance on the `naive' split does not change considerably with our curricular approach.
Results are in Table \ref{tab: Efrat Comparison}, and further training details are given in Appendix \ref{app: efrat}.

\section{Conclusion}

In this paper we introduce a dataset of cryptic crossword clues that can be used as a challenging new benchmark task and 
develop a novel curricular approach that considerably improves performance on the benchmark.
We further characterize the problem with three non-neural baselines and
provide methods for investigating model behavior, including a simple descrambling task and an experiment that explores what T5 learns about compositional task structure.
Lastly we introduce a challenging word-initial datasplit to evaluate a model's ability to achieve compositional generalization.
These contributions demonstrate why this task is worth studying and how it may be relevant to related problems.
For example, our curricular approach may be useful in other settings where focused linguistic capabilities are required.

Pretrained contextual embedding models like T5, which draw on a rich base of lexical and linguistic knowledge from pretraining, are a promising candidate for the type of flexibility needed to solve this sort of puzzle.
However, T5 does not initially succeed at the task, and although our curricular approach considerably improves task performance, cryptics remain an unsolved problem. 
Although one might initially think that a character-level tokenization scheme would be necessary for this task, Sections \ref{analysis:length} and \ref{analysis: descramble} suggest that T5 \textit{can} unscramble words (under appropriate generalization splits) and \textit{does} learn a correspondence between a word's tokens and the word's total length.

Given the success of our curricular approach, future work might combine new synthetic datasets under a learned curriculum schedule.
In any case, an approach that fully solves this problem will need to more flexibly learn different kinds of wordplay functions and how to functionally compose them to produce the answer word. 
In that sense, we believe that the cryptic crossword task serves as a good benchmark for those interested in building NLP systems that can apply linguistic and meta-linguistic knowledge in more creative, flexible, and human-like ways.

\ack

This work was supported in part by a Stanford HAI Hoffman--Yee grant.

We thank Saul Pwanson for guidance on the release of crossword-related datasets and William Tunstall-Pedoe for discussions on cryptics and for testing his proprietary solver on our dataset. 
We also thank Armando Solar-Lezama, Josh Tenenbaum, Osbert Bastani, and other members of the Neurosym group for helpful feedback.

\newpage
\bibliography{acl2020,anthology}

\newpage

\appendix

\section{Clue examples from dataset} \label{app:clue_examples}
\begin{table*}[h]
\small 
\centering
\caption{Examples from our dataset, taken from the train portion of the naive split. Replicates Table~1 in the main paper. Indicators, where they occur, are italicized. The wordplay substrate appears in bold. Typographical emphasis added to aid reader, only; actual clues have no such indication.}
\begin{tabular}{p{0.35\linewidth} @{\hspace{20pt}}  p{0.2\linewidth} @{\hspace{20pt}} p{0.33\linewidth}}
\toprule
  \textbf{Clue type}  & \textbf{Clue example} & \textbf{Explanation for this example}\\
  \midrule
  \textbf{Anagram}: An anagram clue requires scrambling some set of clue letters to produce the answer. & 
  Honour, \textbf{Ben} and \textbf{Noel} \textit{with a new order} (4) &
  \textit{with a new order} indicates that we should re-order (anagram) the letters of ``Ben'' and ``Noel'' to get ENNOBLE. 
   \\ \midrule
  \textbf{Initialism}: An initialism requires taking the first letters of a phrase & 
  \textit{Initially}, \textbf{i}s \textbf{d}octor \textbf{e}lated \textbf{a}t result of brain operation (4) &
  \textit{initially} indicates taking the first letters of ``is doctor elated at'', which gives IDEA, the ``result of brain operation''.
  \\ \midrule
  \textbf{Hidden}: The answer occurs within a larger phrase.  & 
  Cryptic advice f\textbf{or a cl}ever \textit{solver to extract} (6) &
  \textit{solver to extract} indicates that a word is hidden inside a phrase.
  Extract the word ORACLE from the phrase ``for a clever''.
  \\ \midrule
  \textbf{Charade}: For a charade clue, each part of the answer is clued sequentially. & 
  Nitrogen and oxygen shown to exist to student chemist (5) &
  ``Nitrogen'' becomes ``N'', ``oxygen'' becomes ``O'', ``shown to exist'' becomes ''BE'' since  they are synonyms, and a standard abbreviation for student is ``L'' for learner.
  NOBEL was a chemist!
  This clue type does not have an indicator.
  \\\midrule
 \textbf{Double definition}: In a double definition clue, two synonyms or phrases appear next to each other, each of which can refer to the answer. & 
 Painful withdrawal, having raw meat (4,6) &
 ``COLD TURKEY'' means both ``Painful withdrawal'' and ``raw meat''.
 Double definitions do not have indicators.\\ 
 \bottomrule 
\end{tabular}
\label{tab:app_clue_ex}
\end{table*}

\section{Cryptics dataset preprocessing} \label{app:dataset}
To produce the clean dataset, we remove 15,591 clues that interact with other clues in the same puzzle as follows:
\begin{enumerate}
    \item  7,687 clues that are explicitly marked as being part of a clue grouping (i.e. clues that the puzzle author has explicitly marked as interacting).
    For example, from Guardian puzzle 21633:\footnote{Each puzzle can be accessed at \url{https://www.theguardian.com/crosswords/cryptic/puzzle_id}.}
    \begin{enumerate}
        \item  20-across: \inputoutput{this cast no blight on semi-conventional party (8,8)}\\
        \answer{Scottish National}
        \item 5-down: \inputoutput{see 20}
    \end{enumerate}
    In this case, the answer must be written into two locations (20-across and-5 down).
    The first part (20-across) is a valid clue for our models, but we omit clues of this type because programatically parsing them would require simultaneously looking at multiple clues during preprocessing.
    \item 607 ``continuation'' clues or clues that are part of an annotated grouping: 
    These include clues that start with an asterisk (indicating clue grouping) or those that start with an ellipsis, which indicates  continuation from a previous clue.
    For example, from the same puzzle:
    \begin{enumerate}
        \item 23-across: \inputoutput{drunken kilty whams a dram ... (4,6)} \\
        \answer{malt whiskey}
        \item 24-across: \inputoutput{..and another, by the sound of it, on a 20 isle, while ... (4)} \\
        \answer{rhum}
    \end{enumerate}
    Solving 24-across requires having seen 23-across.
    (``Rum'' is a type of malt whiskey that sounds like Rhum, which is a Scottish isle.)
    Note that we also needed to substitute Scottish for 20, from 20-across.)
    \item 7,066 clues that contain a numeral. Many clues with a numeral are references to a solution in another part of the puzzle (i.e. the other solution must be substituted for the numeral). 
    Some numerals are not references, but distinguishing them programatically is not straightforward, so we omit them.
    See for example, the substitution required above of ``Scottish'' for ``20''.
    \item 90 clues that do not match our regular expression or with an empty clue after regular expression extraction.
    \item 56 where the answer does not match the length enumeration. 
    \item 85 where there are unrecognized characters in the clue (e.g., unparsed HTML).
\end{enumerate}

We further remove 1,611 exact duplicates.
These are clues with the same target answer and clue strings that match after lower-casing, normalizing whitespace, normalizing articles (``a'', ``an'', ``the''), and stripping off punctuation. 

In addition to releasing the full dataset, code to fully replicate our data download and pre-processing pipeline is also available in the GitHub repository.
The code we provide reproduces this detailed information, including removal counts broken down by reason, whenever it is run to generate the data splits; comments in the code provide additional details.

\section{Baseline experiment details} \label{app:exp}
We provide details of model and task set-up, hyperparameter choice, machines and compute used, and evaluation methods.

Evaluation is the same for all models. 
When evaluating the correctness of outputs, we lowercase all letters and ignore whitespace.
Generated whitespace (i.e., spaces between generated multiword answers) is considered only for evaluating the meta-properties (e.g., number of words) for model outputs. 
The GitHub repository includes code to exactly replicate all evaluations.

\subsection{WordNet}
The WordNet heuristic approach produces candidate outputs as follows:
the first and last words of a clue are extracted from the clue and lowercased.
For each of these two words, we do a reverse dictionary lookup using WordNet. 
We try building the reverse lookup with synonyms, hyponyms, and hypernyms, where the last two have controllable lookup depth (e.g., hypernyms of the first set of hypernyms, etc).
Any underscores or hyphens in WordNet lookup results are replaced with spaces.
We test with and without inflection of lookup results by using \citep{lemminflect} to produce all possible inflections.
We filter to outputs of the correct length, excluding whitespace. 
We try ranking outputs 
(1) by by their multiset character-level overlap with the rest of the clue (i.e. not the word used for the reverse lookup), 
(2) by bigram overlap with the rest of the clue using a modified Levenshtein distance, 
and (3) by the order in which they are added to the output set (i.e., without further ranking).
For this model, the number of generated outputs is determined by changing which parts of the WordNet graph (synonyms, hyponyms, hypernyms, and depth) we use to generate candidates. 

This model does not involve any training, so the train set is not used.
We take the configuration that produces the best performance on the dev set:
we use reverse lookup with synonyms and hyponyms to depth~1, omit inflected forms, and rank using multiset character-level overlap.

We can upper-bound this method by observing that, when including synonyms and hyponyms/hypernyms up to depth three, and inflecting all outputs using LemmInflect \citep{lemminflect} (i.e., producing the maximum number of candidates for each clue),
our definition sets contains the correct answer 22\% of the time.
This performance could be achieved if we had a perfect ranking function.
However, since our ranking mechanism is poor, we do not achieve this level of performance and find that the best outcome is achieved by reducing the size of our reverse dictionary space to include only synonyms and hyponyms to depth~1.

The slowest of these models is the one with full hyponym/hypernym lookup to depth~3 and was run on a 2013 Macbook Air in two minutes.

\subsection{KNN BoW}
The KNN model is implemented with scikit-learn's \citep{scikit-learn} CountVectorizer and KNeighborsClassifier.
The CountVectorizer lowercases all characters and considers only alphabetic characters, numbers, parentheses and the $\mid$ character. 
All other characters function as split locations and are themselves omitted. 
When including the length enumeration we append length as, e.g., \inputoutput{(4)} or \inputoutput{(4$\mid$6)}, in the case of multiword solutions.
We use `\inputoutput{$\mid$}' so that the length enumeration is treated as a single token. 
As for all other traininable models, targets for the train set are lowercase solutions with spaces separating multiword answers.
We select the 3000 nearest neighbors for each test clue so that we always produce at least ten outputs of the correct length for each clue.

We train by fitting the train set and take the set of hyperparameters that produces the best performance on the dev set:
in particular, we use 1-grams, since performance degrades with longer n-grams. 

This model was run on a 2013 Macbook Air in roughly ten minutes.

\subsection{Rule-based}
We run the \citet{deits_python} solver on our clue sets.
The model is not trainable, so we directly evaluate it on our dev and test sets.
We follow Deits' guidance to set up our clue file, providing a text file where each line is of the form, \inputoutput{clue \pipe{} answer} -- for example,
\begin{center}
\inputoutput{But everything's really trivial initially for a transformer model (4) \pipe{} bert}
\end{center}
We do not restrict the number of outputs generated by this model. 

The rule-based solver uses a context free grammar (CFG) that specifies possible clue forms.
For example, a grammar for an anagram clue type could be “\$Anagram \$AnagramIndicator \$Definition”. 
Terminals for \$AnagramIndicators (and other types of indicators in the full grammar) come from custom lists of indicators. 
One of the components of the CFG is a definition: the definition terminal is matched to a word or set of words. 
The non-definition part of the grammar (“\$Anagram \$AnagramIndicator” in the above example) is evaluated to produce possible wordplay outcomes (in this case, computing valid anagrams of the tokens matched to the \$Anagram terminal).
Finally, the possible wordplay outputs are compared to the definitional tokens using WordNet’s word similarity function. 
Parses with higher similarity are ranked higher.

As mentioned in the footnote in Section~\ref{baseline:rule-based}, \citet{deits_python} has a more recently implemented solver that is reportedly faster.
Because the Python solver is slow, we set a timeout of 120 seconds (timed-out runs usually still produce some candidate answers) and report an average time to solve a clue of roughly 40 seconds.
This results in a total time to evaluate each of the dev and test sets of approximately 300 CPU hours.
We evaluate this model using multiple internal cluster CPUs run in parallel.

\subsection{T5: vanilla seq2seq} \label{app: base:t5}
Starting from HuggingFace's \citep{wolf-etal-2020-transformers} pretrained model parameters, we fine-tune T5-base to produce the target answer for each clue input.
As described in Section~\ref{sec: data_task}, inputs are of the form, e.g., \inputoutput{But everything's really trivial, initially, for a transformer model (4)}, 
with output \inputoutput{bert}.

We optimize with Adafactor \citep{shazeer2018adafactor} using the relative step and warmup initialization options,
as implemented in the HuggingFace library (all other parameters are left unchanged from the HuggingFace defaults).
We use a batch size of 256 input--output (clue--answer) pairs with per-batch truncation-to-longest, which is implemented by HuggingFace's T5FastTokenizer.
We train with a patience of 15 epochs and select the best model according to dev set performance, based on whether the top answer (over 5 beam search generations) is correct.
During evaluation, we generate 100 outputs for each input (100 beams with 100 output sequences) in order to evaluate sample metrics.
Hyperparameters, including those for generation (max-length=10 tokens, length-penalty=0.05), were selected based on dev set performance.
This setup, including all hyperparameters, is implemented in the code that we release on GitHub. 

We use an internal cluster.
Training takes approximately 100 minutes on a single GeForce RTX 3090 GPU.
Evaluation takes roughly 120 minutes.

\section{Curriculum learning} \label{app:curr}

\subsection{Datasets} \label{app: curr: datasets}

\subsubsection{ACW-data} \label{app: curr: datasets: acw}
ACW-data is the unprocessed version of the American crossword clue dataset \citep{xd_cw_paul}.
To preprocess it, we 
\begin{enumerate}
    \item Remove clues that do not match our reverse-dictionary goal:
    We remove clues that contain underscores or multiple hyphens, since these are generally fill-in type clues, rather than phrasal synonyms. 
    We remove clues that reference other clues, i.e., those containing ``Across'' or ``Down'' in the clue text.
    We remove clues likely to be abbreviations, i.e., those with a clue ending in a period with an answer fewer than four letters, since cryptics rarely include abbreviations.
    We remove clues where the clue text ends in a question mark.
    \item We attempt to make the clues resemble our dataset by removing any periods that occur at the end of clues, since cryptic clues do not generally have periods at the end of normal clues (though they do admit other types of punctuation).
    \item We filter normalized duplicates using the same approach as for cryptic clues (i.e. clues with the same clue and answer strings after normalizing case, whitespace, and articles and stripping punctuation.
\end{enumerate}
This produces a cleaned dataset of 2,464,951 clue-answer pairs from which we produce the three ACW-data-derived datasets used in curricular training. 
It is worth noting that some of the answers in this dataset are multiword answers that are unsplit.
Optimally we would find a way to split these answers to increase similarity to our primary dataset, which does split multiple word targets.

The code to reproduce this preprocessing and to produce the following datasets is included in the GitHub repository.
Details of the three datasets (ACW, ACW-descramble, and ACW-descramble-word were given in the main paper (Section~\ref{curr: datasets}).

\subsubsection{ACW training datasets}
The actual input-output pairs for ACW, ACW-descramble, and ACW-descramble-word are produced from the processed version of ACW-data at train time.
At train time, we prepend a task label as described in the main text.
The ACW curricular dataset has no further modification.
For ACW-descramble and ACW-descramble-word, we produce a scrambled version of the letters during dataset collation and modify the input as specified in the main text.
The provided code includes the collation functions that produce the final input--output pairs for these three datasets.

\subsubsection{Anagrams dataset}
First, we produce a list of valid English words to be considered for anagramming from a publicly available dictionary of English words.
Using this list of words, we group all words into whether they are anagrams of each other (i.e. grouping them by their sorted letters).
For anagram indicators, we use \citet{deits_python} list of anagram indicators.

This produces 13,535 anagram groups (i.e., 13,535 unique sets of letters from which can be realized at least two valid English words).
These groups contain a total of 32,305 total words.
The anagram indicator list has 1,160 anagram indicators.
At train time, a curricular epoch consists of showing each anagram group to the model once.
To do this, during collation at train time, we randomly sample two of the anagrams from each set, randomly sample an anagram indicator, and randomly sample a position (prepend or append).

\subsection{Training} \label{app: curr: training}
As described in Section~\ref{curr: datasets}, each supplementary dataset has its own task label \citep{2020t5}, which is passed to the model as part of the input string, and all inputs include length enumeration as in the vanilla T5 case.
We fine-tune T5-base in the same way as described in Appendix~\ref{app: base:t5}, but with the following modifications.

For curricular training, we first fine-tune on one or more supplementary tasks according to a training schedule, for which we tune the following hyperparameters: the number of curricular epochs, the frequencies with which each task is shown, whether the Adafactor optimizer is reset before primary training (only affects non-constant LR, i.e., when we are training T5-base but not when we are training T5-large), and the frequency of curricular subtask review during primary training. 
We hand-tune these hyperparameters, generally finding that training nearly to convergence on a held-out dev set for the primary curricular task is optimal. 
We also find that, for T5-base, resetting the optimizer between curricular and main training slightly improves performance.
The specific configurations to replicate curricular training are included in the GitHub repository.

In order to directly compare the different curricula, we set up the curricula so that the number of training examples shown to the model in each epoch as well as the mix between curricular and primary task are the same. 
For example, for our single-dataset curricula (ACW and ACW-descramble), we run experiments with 4 curricular epochs and relative batch frequences (primary dataset: curricular dataset) during main training of 20:6. 
When training on curricula that include two curricular datasets, we do only 2 curricular epochs and use relative batch frequencies of 20:3:3 (primary: curricular 1: curricular 2).

To produce Table~{\ref{tab: curricular}}, we evaluate only on the dev set over five generations to enable faster iteration.
To produce the second column of the table, we algorithmically identify anagram clues.
Code to replicate the anagram labeling and evaluate on this subset is available in the GitHub repository.

To produce our top result in Table~\ref{tab: Results_main}, we double the total number of curricular epochs (from 2 to 4), select the best model checkpoint via dev set performance, and perform final evaluation on the test set taking 100 generations.

For all curricular training we use an internal cluster.
Each curricular epoch takes roughly 150 minutes, giving a total curricular training time of roughly ten hours.
Primary training afterward takes roughly 130 minutes since we continue to review the curricular datasets.
This gives a total train time of roughly 12 hours on a single GeForce RTX 3090 GPU.

\section{Model analysis details}

\subsection{Descrambling task} \label{app: exp: descramble}

We start with the preprocessed version of ACW-data from Appendix~\ref{app: curr: datasets: acw} and further remove any clue--answer pair with an answer that is not in an English dictionary (e.g., multiword answers would be removed).
This guarantees that all descrambling targets are valid English words.

After removing multiword answers, we have a dataset of 1,796,078 clues.
We keep only words that have between 4 and 14 characters and downsample to 10\% (roughly 180k clue-answer pairs).

We train T5-base to complete the descrambling tasks using the same approach as in Appendix~\ref{app: base:t5}.
Code to replicate dataset creation, training, and evaluation are available in the GitHub repository.

\subsection{Wordplay systematic learning} \label{app: wordplay}
Detailed code that identifies first name anagram substrates and generates substitutions is included in the GitHub repository.
For name identification, we use names lists from the US Naval Academy and the U.K. Office of National Statistics (both lists, including with download URLs are provided in the GitHub repository).
We identify 27 clues in the dev set and 69 clues in the train set that require anagramming a single word that is also a first name, and for each we perform 10 scramble and 10 name substitutions.

\subsection{Efrat et al training} \label{app: efrat}
We use the same training setup as in Appendix~\ref{app: base:t5}, but with the following changes:
we train T5-large with a constant learning rate of 3e-5 and an effective batch size of 768.
For evaluation we use the same metric (top output with b=5 beams, no filter) as used by \citet{efrat2021cryptonite}.

We again train on an internal cluster using a single GeForce RTX 3090 GPU.
Training to replicate \citet{efrat2021cryptonite} results (i.e. non-curricular) takes roughly ten hours.
Curricular pretraining is done for 3 epochs and takes roughly 4 hours per curricular epoch, giving a total time for curricular pretraining of roughly 12 hours.

Code to replicate this approach is included in the GitHub repository.

\end{document}